# Can Public LLMs be used for Self-Diagnosis of Medical Conditions?

Nikil Sharan Prabahar Balasubramanian*[1]
nbalasubramanian@uttyler.edu

Sagnik Dakshit*[1]
sdakshit@uttyler.edu

[1]Computer Science, The University of Texas at Tyler, TX, USA
*Authors have equal contribution

## Abstract

Advancements in deep learning have generated a large-scale interest in the development of foundational deep learning models. The development of Large Language Models (LLM) has evolved as a transformative paradigm in conversational tasks, which has led to its integration and extension even in the critical domain of healthcare. With LLMs becoming widely popular and their public access through open-source models and integration with other applications, there is a need to investigate their potential and limitations. One such crucial task where LLMs are applied but require a deeper understanding is that of self-diagnosis of medical conditions based on bias-validating symptoms in the interest of public health. The widespread integration of Gemini with Google search and GPT-4.0 with Bing search has led to a shift in the trend of self-diagnosis using search engines to conversational LLM models. Owing to the critical nature of the task, it is prudent to investigate and understand the potential and limitations of public LLMs in the task of self-diagnosis. In this study, we prepare a prompt engineered dataset of 10000 samples and test the performance on the general task of self-diagnosis. We compared the performance of both the state-of-the-art GPT-4.0 and the fee Gemini model on the task of self-diagnosis and recorded contrasting accuracies of 63.07% and 6.01%, respectively. We also discuss the challenges, limitations, and potential of both Gemini and GPT-4.0 for the task of self-diagnosis to facilitate future research and towards the broader impact of general public knowledge. Furthermore, we demonstrate the potential and improvement in performance for the task of self-diagnosis using Retrieval Augmented Generation.

## 1. Introduction

Deep learning has emerged as a dominant paradigm in the field of Natural Language Processing (NLP), revolutionizing the way we interact and understand textual data such as conversations. From language translation and sentiment analysis to text generation and summarization, deep-learning techniques have demonstrated unparallel performance across a wide range of NLP tasks, surpassing traditional rule-based and statistical methods. The success of deep learning in NLP can be attributed to the large-scale availability of natural language data and the ability of deep learning to learn intricate patterns and representations directly from raw data, without the need for handcrafted features. Despite the success of deep learning models, their lack of interpretability and transparency makes it difficult to understand the inner workings of these complex systems and the reasoning for their outputs. Unlike traditional machine-learning models, which provide explicit explanations for their predictions based on interpretable features or rules, deep-learning models operate as black boxes, making it challenging to understand how they arrive at decisions. This problem is exacerbated by the widespread use of deep learning models by the general public, with increasing accessibility to Large Language Models such as ChatGPT, and their integration in various software and applications.

Large Language Models (LLMs) have emerged as transformative tools in the field of natural language processing (NLP), demonstrating remarkable capabilities for understanding and generating human-like text. These models, fueled by advances in deep learning and trained on large amounts of data, have revolutionized various NLP tasks, including but not limited to translation, text summarization, and question answering, and have raised questions about their potential as Artificial Generalized Intelligence (AGI). While early iterations of LLMs were primarily proprietary and accessible only to a few organizations with extensive computational resources, the advent of open-source access has led to publicly available state-of-the-art LLMs. Models such as OpenAI's GPT series, Google's Gemini, Meta's Llama and Minstral have been made accessible to researchers, developers, and enthusiasts worldwide, ushering in a new era of NLP research and application development for the public. One such domain of rapid growth and adoption of deep learning models and LLMs for various task automation is in healthcare, with the goal of improving patient outcomes and healthcare services provided by burdened healthcare professionals.

The rise of digital health technologies and the widespread availability of health information on the internet have empowered individuals to take a more active role in managing their health and well-being. One notable phenomenon observed in this trend is the emergence of medical self-diagnosis, in which individuals use online resources, symptom checkers, and mobile applications to assess their own health conditions without consulting a healthcare professional. While not recommended, a staggering 70% [1] of the American population was reported to have consulted the internet for medical reasons. While medical self-diagnosis offers the potential for greater autonomy, convenience, and accessibility in healthcare decision making, it also raises significant concerns regarding accuracy, reliability, and safety. With the shift in paradigm from long-standing Google searches to LLMs such as GPT, it is prudent to study the reliability and trustworthiness of the medical diagnosis of these models for general population awareness. Often, these searches are based on patient bias, leading to misinformation [2]. The lack of explanations in deep learning models coupled with the widespread use of Large Language Models based on black-box deep learning models and bias-validating searches poses significant challenges that are of significant importance and in the best interest of general public health, where model interpretability is critical for ensuring transparency, accountability, and trustworthiness.

In this paper, we investigate the potential of public LLMs for the task of self-diagnosis of medical conditions to explore the reliability and trustworthiness of the decisions. To the best of the author's knowledge, this is the first work on investigation of self-diagnosis using both free and public LLM models with over 10000 medical searches in the interest of general public health and awareness. Our contributions in this paper can highlighted as follows:
- Investigation of self-diagnosis of medical conditions with 10000 samples on state-of-the-art free LLM models available to the general public, namely Gemini and GPT-4.0.
- We discuss the challenges and potential of LLMs in self-diagnosis.
- We demonstrated the importance of the user role that aids the LLM's learned knowledge and bias in making predictions.
- We demonstrate that the potential of LLM in self-diagnosis lies in its use through Retrieval Augmented Generation (RAG), which leads to a significant improvement in performance, with 98% accuracy.
- We also discuss the challenges of processing existing datasets that can be modeled for the task of self-diagnosis for future researchers as guidance to facilitate even larger-scale studies that are important for general public health and moderators.

## 2. Related Works

This section discusses relevant research on the application of LLM for medical diagnosis in general, and specifically on the task of self-diagnosis. Owing to the nascent wide-spread development of LLMs and lack of conversational datasets for self-diagnosis, this important task of critical public health has been explored in limited capacity. Koga et. al [3] discussed studies on the limitations of ChatGPT in medical diagnosis highlighting the risk of misinformation. Kuroiwa et al. [4] evaluated the performance of ChatGPT under five common orthopedic conditions. The authors obtained a 66.4% correct answer ratio on the five questions asked daily for a period of five days, making it a small-scale evaluation, and observed inconsistency in its answers to the same question. In addition, Kuroiwa et al.. suggested that it is important to highlight and recommend medical consultations. Similar inconsistencies were observed in the answers to pathology exam questions [5] and random prompts [6]. In \cite{ten2024chatgpt}, authors investigated the ability of ChatGPTv3.5 in differential diagnoses of patients based on physician notes recorded at initial ED (Emergency Department) presentations. The authors observed comparable physician and ChatGPTv3.5 performance which significantly improved over physician accuracy with additional lab results along with clinician's notes. The authors concluded that ChatGPTv3.5 can aid as clinician's aid but not as a replacement. Authors Hirosawa et. al [8] investigated both ChatGPTv3.5 and ChatGPT-4 on case vignettes from 52 case reports published by the Department of GIM of Dokkyo Medical University Hospital, Japan and recorded 83%, 81%, and 60% accuracy for 10 differential diagnosis lists, top 5 differential diagnosis lists, and top diagnosis respectively. However, these studies did not investigate non-clinical self-diagnosis searches, which have a larger societal impact and are of significant importance to public health.

The largest study on this topic involving 500 medical prompts on diabetes and migraines was used, and the results from BERT and GPT-4 were evaluated by a panel of medical experts with 80% accuracy on specialized medical queries and compared to answers from a group of medical students [9]. These performances were achieved by fine-tuning the BERT and GPT-4 models which is not feasible for general public hindering the relevance of the results this study to public health. In [10], the authors developed and refined LLM on their developed patient-doctor conversational dataset. By leveraging linear transformations, conversational chest X-ray diagnostic tools were developed in [11]. Extending to other languages, researchers have fine-tuned LLaMa [12] to Chinese medical knowledge [13]. Li et al. [14] proposed BEHRT, which can predict the most likely disease of a patient in their next visit by learning from EHRs, while Med-BERT [15] predicts heart failure in diabetic patients.

An important study in the investigation of LLM for self-diagnosis modified the United States medical board exams, which were modified as self-diagnostic reports from patients [2]. The discussion and results highlight that for bias-validating information, which is common in self-diagnosis, the performance drops significantly. These studies do not investigate publicly available models which are more likely to be used by general population and are also limited in terms of the data size of investigation. Furthermore, the general public does not have a detailed or accurate representation of symptoms or diagnosis, such as doctors' notes or laboratories, which increases the associated risk and the critical nature of the self-diagnosis task. To the best of the author's knowledge, our work is the first large-scale investigation with 10000 prompts of self-diagnosis nature without clinical details or lab reports on two state-of-the-art public LLM models namely GPT-4 and Gemini that are now available freely through their integration with Google and Bing search.

## 3. Methodology

In this section, we discuss our design methodology for formatting data, model selection, processing and evaluation of the task of self-diagnosis. Self-diagnosis differs from conversational diagnosis such as observed in a clinical setting. While conversational diagnostic involves the expert asking a series of questions and answers for validation of symptoms which eventually lead to a condition identification, in contrast self-diagnosis is bias-validating through a single question with a list of symptoms being experienced or being perceived. The nature of self-diagnosis as can be thought of in a google search limits the availability of suitable public datasets. Our pipeline transforms the conversational dataset into a self-diagnosis dataset, following which we undertake prompt engineering with added context and restrictions to pass to LLM model. Furthermore, for comparison of LLM models, we ensure similar formatting of results by imposing restrictions through prompt and also processing the outputs for performance evaluation.

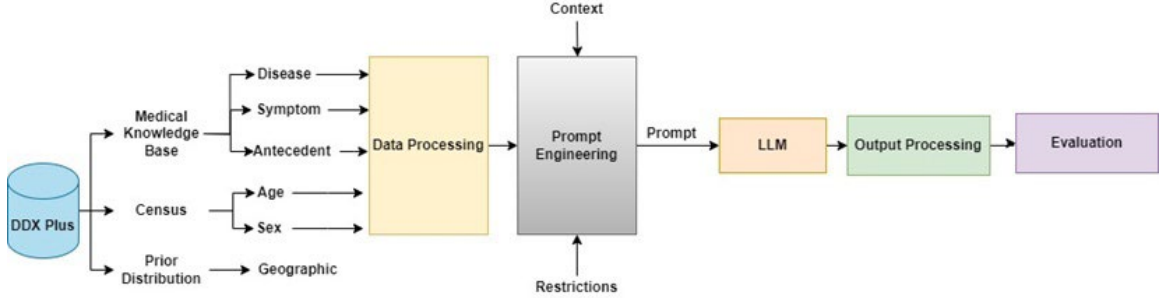

Fig. 1: Research methodology pipeline with DDXPlus Data processing, prompt engineering, and Output Processing modules for evaluation of public LLM performance on the task of self-diagnosis of medical conditions.

**3.1 Model Selection**

We selected two public LLMs, Gemini and GPT-4.0, for our investigation in this paper. Our selection of GPT-4.0 is based on a literature survey demonstrating its high potential for various tasks, including medicine [16], education [17, 18], drug event detection [19], finance [20], and law [21]. Despite being available for free use through a Bing search, the GPT-4.0 API is used for extensive testing and research tasks. To compare, we select Gemini which is a free model and accessible to a wider general population through Google search. Rane et al. in their work [22] compared Gemini with GPT and discuss Gemini's factual accuracy over GPT's conversational supremacy. Buscemi et. al [23] also included Llama in their comparison of multilingual sentiment analysis and demonstrated comparable performance from all three LLM families with certain biases in Llama.

**3.2 Dataset and Processing**

In this study, we used one of the largest medical diagnosis datasets, DDX Plus [24], which contains patient census data, namely age and sex, geographic locations, and a medical knowledge base consisting of diseases as conditions, symptoms as evidence, and antecedents. Symptoms are specific indicators or manifestations associated with a condition. In the DDXPlus dataset, each symptom was identified by a unique code, such as $E\_65$, $E\_63$, $E\_52$, $E\_172$, $E\_84$, $E\_66$, $E\_90$, $E\_38$, and $E\_176$ for Myasthenia gravis. These codes help in systematically cataloging and analyzing the symptoms observed in a patient. Antecedents are previous medical conditions or factors relevant to the patient's current health status, which are also coded with identifiers, such as $E\_28$ and $E\_204$, representing specific historical health factors of the patient and evidence as critical in establishing the presence of the condition and can vary in type. For example, binary evidence is simply noted by its name (e.g., "$E\_28$"), while categorical or multi-choice evidences are represented in a format like evidence-name_@_evidence-value (e.g., smoking_@_heavy. Multiple entries for multi-choice evidences can be present if different values are applicable. There is a total of 223 evidences, 110 symptoms, and 113 antecedents, as illustrated in Fig. 2, and the permutation and combinations resulting in 1025602 synthetic patients for our purpose of self-diagnosis investigation using public LLMs. We do not include any geographic information to minimize bias in our study. Running each data sample as a prompt along with their User Role and System Role as illustrated in Section 3.3, costs $0.005 on GPT-4.0, which limits the scope of our investigation, as this is an unfunded project. We randomly select 10000 data samples for our investigation on GPT-4.0 and maintain the same for comparison with free Gemini LLM Model.

**3.3 Prompt Engineering**

The structure of the dataset does not allow direct investigation of self-diagnosis as the samples in the dataset are posed for conversational diagnostics in a clinical setting and not for self-diagnosis, as illustrated above. Furthermore, the consistency and quality of results obtained from LLMs are directly dependent on

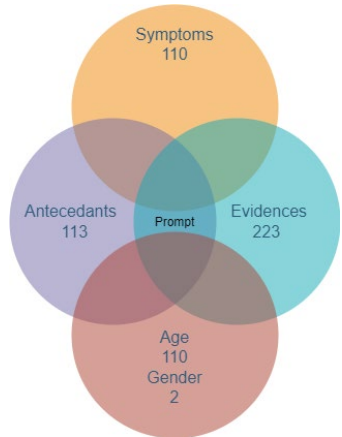

Fig.2: Prompt engineering using DDXPlus data. The 110 symptoms, 113 antecedents, 223 evidences are combined with census data of age and genders to develop 1025602 synthetic patients

the prompts, which has led to the development of studies on prompt engineering. We combined evidence, symptoms, antecedents, and census data to generate our prompt, as shown in Fig. 2. Studies have demonstrated that context provided through the User Role in a prompt not only helps LLMs understand the intent and produce a better response, but also consistent responses denoted as UR in Fig. 1. To ensure consistency in the output pattern, we also set some restrictions through the System Role for the LLM, which are represented as the SR in Fig. 1. We use an empirically established fixed leading system role along with the age and gender from the census data as our user role to develop our prompt's leading context. In between the leading and trailing context, we pass the positive evidence and antecedents as our engineered prompt for medical condition identification. For both the LLMs, the following system role (SR) was used to get the desired response:

System Role (SR):
- *You are an AI trained to help with medical diagnosis based on symptoms described by users.*
- *Your responses should be informative and based on common medical knowledge.*
- *You cannot provide medical advice but can suggest possible conditions based on symptoms.*
- *You cannot generate a response to this message.*
- *Only give me a list of diseases and no additional words.*
- *The names of the diseases must be in a single line separated by commas*

Following the leading system role, the following user role is used as part of the prompt for all the requests, where the variables are populated from the census data and the corresponding antecedent, symptoms.

User Role (UR): *I am a {age}-year-old {sex}. I have been asked the following questions about my symptoms and antecedents: {symptoms}. I have answered the questions I am sure about. What is my diagnosis? Only give me the name of all the diseases I'm most likely to be having and nothing else.*

One such example of a prompt with the user role is provided below.

*"I am a 18-year-old Male. I have been asked the following questions about my symptoms and antecedents: Do you live with 4 or more people?, Have you had significantly increased sweating?, Do you have pain somewhere, related to your reason for consulting?, Characterize your pain: - sensitive, Characterize your pain: - heavy, Do you feel pain somewhere? - forehead, Do you feel pain somewhere? - cheek(R), Do you feel pain somewhere? - temple(L), How intense is the pain?, Does the pain radiate to another location? - nowhere, How precisely is the pain located?, How fast did the pain appear?, Do you have a cough that produces colored or more abundant sputum than usual?, Do you smoke cigarettes?, Do you have a fever (either felt or measured with a thermometer)?, Do you have a sore throat?, Do you have a cough?, Have you traveled out*

*of the country in the last 4 weeks? - N, Are you exposed to secondhand cigarette smoke on a daily basis? I have answered the questions I am sure about. What is my diagnosis? Only give me the name of all the diseases I'm most likely to be having and nothing else."*

As can be observed from the example, for some of the data samples, all the symptomatic questions have not been answered which can have a significant impact on the results of these LLM Models. The system and the user roles are mandatory requirements for use of GPT API thus limiting our scope of experimentation on the results without specific user and system roles.

### 3.4 Output Processing and Evaluation

Owing to our imposed restrictions on the LLMs, passed as prompt and illustrated in Section 3.3, the LLMs produce an output as a list of potential medical conditions. If any of the conditions listed matched the true diagnosed conditions in the dataset, we consider it to be correct prediction by the model for our use case. Iterating over all our 10000 engineered prompt samples, we calculate the accuracy of the LLM models.

We processed the LLM outputs through our Output Processing Module to compare the LLM-predicted diseases against the actual conditions to calculate accuracy as a measure of performance evaluation through string matching. This analysis allowed for an efficient evaluation of GPT's predictive accuracy of the GPT based on the prompts provided. The processing of Gemini required additional formatting over GPT owing to several inconsistencies, as discussed in detail in Section 5.1. Output processing for Gemini was crucial to accurately handle and correct these issues before performing accuracy calculations.

A challenge common to both GPT and Gemini outputs evolved from the use of disease acronyms in the original dataset, where diseases were labeled with acronyms such as "*URTI*" for *Upper Respiratory Tract Infection*. Both language models sometimes spelled out these diseases in full during their responses, which required a unified solution to map the full disease names back to their corresponding acronyms using dictionaries to maintain consistency with the dataset's format and measure performance through accuracy. Our output processing modules handle the above illustrated challenges for both Gemini and GPT models through specific adjustments for accurate interpretation and performance evaluation. This highlights the limitation and need for adaptable data handling techniques when working with varied outputs from different language models.

## 4.0 Self-Evaluation and Bias Results

Owing to the large cost of using the public LLM GPT-4.0 API, which costs $0.005 per prompt, we run only a randomly selected subset of the dataset with 10000 cases. To the best of the author's knowledge, this is the first attempt in investigating the performance of public LLM models for the task of self-diagnosis of medical conditions. Using our 10000 sample data points, we recorded an accuracy of 63.07% with GPT-4, while Gemini achieved a significantly lower accuracy of 6.01%. The results clearly show that neither considered LLMs have the capacity to provide any guidance on self-diagnosis and can have serious effects from unrestricted usage by the general population.

Through our experimentation, we also test for diagnosis biases that exist in both the considered LLMs and the importance of user role. We investigated the presence of bias by changing the user role, as illustrated in Section 3.3, to the other extreme of the spectrum. For illustration we pass the same prompt by changing the age and sex from 18 years old male to 90 years old female and pass the prompt with the same evidence and antecedents. We observed that the list of possible conditions generated changes significantly and can be illustrated as follows:
- Gemini
  - User Role: 18-year-old male

- Correct: URTI, Predicted: Asthma, Diabetes, Hypertension, Malaria, Typhoid, Epilepsy, Meningitis, Pneumonia, Stroke, HIV/AIDS.
  - User Role: 90-year-old female
    Correct: URTI, Predicted: Trigeminal neuralgia, Cluster headache.
- GPT-4.0
  - User Role: 18-year-old male
    Correct: URTI, Predicted: Sinusitis, Tension headache, Migraine, Upper respiratory infection
  - User Role: 90-year-old female
    Correct: URTI, Predicted: Sinusitis, Migraine, Temporal Arteritis, Upper Respiratory Infection

This change in predictive diagnosis is the sole impact of changing the user role based on census data, demonstrating the presence of bias and the importance of the user role in the output of the considered language models. *"Bias free learning is futile"* has long been established as both a good and evil in terms of model learning and through our experiment, we do not investigate the correctness of the bias due to missing gold standard data but only demonstrate the presence of knowledge and bias in the LLM.

## 5.0 Discussion on Challenges and Potential

### 5.1 Challenges

The widespread application and integration of LLMs in research and everyday critical tasks such as self-diagnosis prompts the requirement of understanding their reliability, potential and technical challenges for future research. To this end, we grouped the identified challenges into three categories: (1) Technical challenges, (2) Formatting challenges, and (3) Inconsistency challenges. In this section, we discuss the above challenges in context of both Gemini and GPT-4.0 models.

- The technical reliability of LLMs is crucial in research settings. It's not only about the accuracy of the content, but also about delivering it in a practical format and maintaining system stability throughout the research process. In contrast to GPT-4.0, significant technical challenges were faced with Gemini which hindered the usability in large scale testing exacerbated by the lack of documentation. The Gemini API had frequent crash error listed as "other" and required restarting the system.
- Formatting Challenges: While GPT-4 adhered to the prompt instructions, the responses for Gemini were inconsistent in formatting and required significant manual reformatting for accuracy analysis, as illustrated in Section 3.4 and are presented in Figure 3. Gemini's output formatting did not align with our prompt instructions in about 20% of the cases, but it also required manual reformatting. Gemini's multiple crashes and inconsistent formatting demanded a notable amount of extra time and attention, underscoring the importance of using robust and dependable LLMs like GPT-4 that streamline the research process by eliminating unnecessary manual intervention.
- Inconsistent Disease Predictions: One of the main challenges observed was inconsistent predictions over multiple runs using the same prompt. This problem was exacerbated for Gemini LLM in contrast to GPT-4.0, for which the correct condition was always included in the predicted list while some other possible conditions changed. We ignored this in our calculation of accuracies as general population is unlikely to try the same prompt multiple times for medical condition prediction.

### 5.2 Potential

The results demonstrate the potential of GPT-4.0, with significantly superior performance over Gemini, but its limited performance of 60% accuracy calls for stricter restrictions in terms of answers provided to the general population. The performance of GPT-4 proved that it has the potential to be used as a self-diagnostic tool, but a lot of considerations should be made. GPT-4 also had better performance in terms of adhering to

```
Acute pulmonary edema,"* Malaria
* Typhoid
* Dengue
* Bronchitis
* Upper respiratory tract infection (URTI)"
 Inguinal hernia,"- Hypertension
 - Diabetes
 - Asthma
 - Obesity
 - Depression
 - Anxiety"
* **Hepatitis B**
* **Hepatitis C**
* **Hepatitis D**
* **Cirrhosis**
Guillain-Barré syndrome,"**Possible diseases:**
Panic attack,"Age: 38 years
Sex: Female
Symptoms: Tiredness, unintentional weight loss,
Possible Disease: Diabetes mellitus type 1"
```

Fig. 3: Inconsistent Formatting of Gemini Output: (A) Prefix asterisk, missing comma delimiter and not in a list, (B) Prefix hyphen, missing comma delimiter and not in a list, (C) Prefix and postfix asterisk, missing comma delimiter and not in a list, (D) Unwanted text "Possible diseases" (E) Unwanted information age, sex, symptoms and presence of quotes.

prompt instructions over Gemini. The main potential of the LLMs lies in enhanced efficiency and accessibility. For regions with limited access to healthcare professionals, AI-driven tools could provide essential support by offering preliminary assessments that inform patients about the urgency of their condition. Integrating language models like GPT-4.0 into primary care settings can significantly reduce the wait times for initial consultations, helping prioritize more urgent cases and manage patient flow more effectively but the models need to be fine-tuned or developed specifically on clinical data. LLMs could help in triaging symptoms, suggesting potential conditions based on user inputs, and advising on the urgency of seeking medical attention. This potential can significantly improved through fine-tuning on extensive, diverse, and high-quality healthcare-specific datasets as AI models interact with more data and feedback, they can continuously learn and improve their diagnostic predictions. The possibility of their use based on Retrieval Augmented Generation (RAG) models also needs to be investigated. For both RAG and healthcare specific models, effective prompt engineering is crucial to ensure the AI interprets user input accurately and delivers clinically relevant responses. To leverage this potential, high accuracy and reliability are critical for patient safety, requiring ongoing monitoring and validation. Furthermore, stringent data protection measures and compliance with healthcare information privacy laws and must undergo rigorous regulatory scrutiny to meet healthcare standards.

**5.3 Retrieval Augmented Generation Results**

Retrieval Augmented Generation (RAG) is an emerging application paradigm of Large Language Models (LLMs) that seeks to combine the strengths of large language models with domain-specific knowledge sources. As demonstrated above, LLM models, while having potential for medical diagnosis tasks face several challenges in terms of formatting limiting automation and reliability in terms of consistency thus hindering their potential. RAG addresses these limitations through a retrieval mechanism that enables the model to access domain-specific databases or knowledge sources at inference time, thereby enhancing its ability to produce more precise and informative responses.

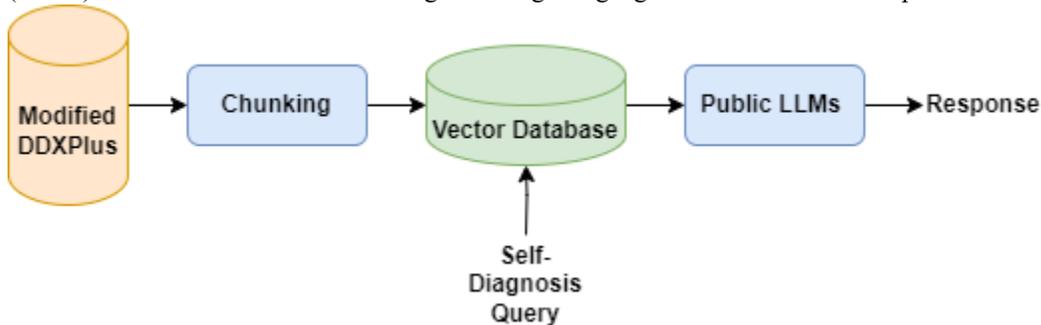

Fig.4: RAG pipeline using public Gemini and GPT-4.0 LLMs developed on processed DDXPlus data as a domain-specific knowledge base.

In this section, we develop a knowledge base for self-diagnosis from our compiled set of symptoms, antecedents, census, evidence and evaluate the potential of RAG model on this domain-specific knowledge base. We investigate with both GPT-4.0, and Gemini based RAG models on the same set of prompts. This cannot be considered as data leakage as we are not training any model but investigating the potential of pre-trained public LLM on knowledge bases which is traditional for RAG models. Both the Gemini and GPT-4.0 RAG models achieved 100% accuracy, which is a significant improvement. However, we noticed that consistency was still a limiting challenge of Gemini RAG. While the correct disease was always predicted, the other potential conditions listed often varied across runs. An interesting observation we recorded was Gemini RAG prompting for additional symptoms, the evidence to which were missing and concluding that a diagnosis was not possible without the specific a specific antecedent and symptom. This demonstrates the potential of RAG using LLMs for the task of self-diagnosis.

## 6.0 Conclusion

The recent advances in deep learning have led to transformative developments in various domains through their applications. In the field of natural language processing, the success of Large Language Models has led to their widespread integration in various applications across domains from education to healthcare. Furthermore, the rapid progress coupled with the open-source services has made it accessible to public. Tasks such as writing, summarizing, and even web searching have observed a paradigm shift with increased leveraging of these LLM models such as GPT, Gemini. However, the reliability and integrity of these models have been questioned in terms of hallucination which poses a significant risk for critical tasks such as self-diagnosis. Self-diagnosis is the task of posing a set of experienced or bias-validating symptoms as a question to a non-medical entity for a diagnosis. Furthermore, this differs from traditional conversational diagnosis in a clinical setting. While the general population has long relied on Google search as a medium for self-diagnosis, recent developments have led to a shift to dependency on LLM for the task. In the interest of public health, it is crucial to investigate the potential and limitations of LLM models in the task of self-diagnosis for developing policies, restricting responses and for facilitating future developments and research. To the best of the author's knowledge this is the first investigation on the use of LLM for self-diagnosis. In this paper, we propose a pipeline to transform a traditional conversational medical diagnosis dataset to fit the task of self-diagnosis and compare both free and public state-of-the-art LLMs, namely GPT-4.0, by OpenAI and Gemini by Google, achieving 6.01% and 63.07% accuracy respectively on 10000 data samples. Furthermore, by changing the user role, we demonstrate its importance in leveraging the LLM's knowledge and bias to make predictions. We also discuss the challenges of consistency, reproducibility of LLM outputs as well as their potential and limitations. As future work, we plan to investigate with larger datasets and a wider range LLMs including non-public models. Furthermore, we would like to investigate the minimal number of symptoms and nature of symptoms which are important to identify medical conditions correctly by the LLMs.